\documentclass[letterpaper, 10 pt, conference]{ieeeconf}  

\IEEEoverridecommandlockouts                              
\usepackage{amsmath} 
\usepackage{amssymb}  
\usepackage[utf8]{inputenc}
\usepackage{flushend}
\usepackage{graphicx}
\usepackage{cite}
\usepackage{siunitx}
\usepackage[caption=false]{subfig}
\usepackage{afterpage}
\usepackage{booktabs}
\usepackage{siunitx}
\usepackage{multicol}
\usepackage[bookmarks=true]{hyperref}
\usepackage{mathdots} 
\usepackage{rotating}
\usepackage{microtype}
\newcommand{\rotHeader}[1]{\begin{sideways}\textbf{#1}\end{sideways}}

\title{SphNet: A Spherical Network for Semantic Pointcloud Segmentation
\author{Lukas Bernreiter, Lionel Ott, Roland Siegwart and Cesar Cadena} 
\thanks{This work was supported as a part of NCCR Robotics, a National Centre of Competence in Research, funded by the Swiss National Science Foundation (grant number 51NF40\_185543).}
\thanks{All authors are with the Autonomous Systems Lab, ETH Zurich, Zurich 8092, Switzerland, {\tt \small \{berlukas, lioott, rsiegwart, cesarc\}@ethz.ch.}}%
}

\usepackage[utf8]{inputenc}
\usepackage[OT1]{fontenc}

\usepackage{fancyhdr}

\usepackage{textcomp}\usepackage{gensymb}

\usepackage{multicol}

\usepackage{subfig}
\usepackage{graphicx}

\usepackage{upgreek}

\usepackage{units}

\usepackage{pdfpages}
\includepdfset{pages={-}, frame=true, pagecommand={\thispagestyle{fancy}}}

\usepackage[nolist,nohyperlinks]{acronym}

\newcommand{\norm}[1]{\left\lVert#1\right\rVert}

\renewcommand{\vec}[1]{\ensuremath{{\boldsymbol{#1}}}}

\begin{acronym}
\acro{SLAM}{Simultaneous Localization And Mapping}
\acro{TSDF}{Truncated Signed Distance Field}
\acro{ICP}{Iterative Closest Point}
\acro{RFS}{Random Finite Sets}
\acro{CNN}{Convolutional Neural Network}
\acro{FoV}{Field-of-View}
\acro{mIoU}{mean Intersection over Union}
\end{acronym}

\begin{document}

\maketitle

\begin{abstract}
Semantic segmentation for robotic systems can enable a wide range of applications, from self-driving cars and augmented reality systems to domestic robots. 
We argue that a spherical representation is a natural one for egocentric pointclouds. 
Thus, in this work, we present a novel framework exploiting such a representation of LiDAR pointclouds for the task of semantic segmentation. 
Our approach is based on a spherical convolutional neural network that can seamlessly handle observations from various sensor systems (\textit{e.g.}, different LiDAR systems) and provides an accurate segmentation of the environment.
We operate in two distinct stages: First, we encode the projected input pointclouds to spherical features. 
Second, we decode and back-project the spherical features to achieve an accurate semantic segmentation of the pointcloud.
We evaluate our method with respect to state-of-the-art projection-based semantic segmentation approaches using well-known public datasets. 
We demonstrate that the spherical representation enables us to provide more accurate segmentation and to have a better generalization to sensors with different field-of-view and number of beams than what was seen during training.
\end{abstract}

%
%
\section{Introduction}
\label{sec:Introduction}
Over the past years, there has been a growing demand in robotics and self-driving cars for reliable semantic segmentation of the environment, \textit{i.e.,} associating a class or label with each measurement sample for a given input modality.
A semantic understanding of the surroundings is a critical aspect of robot autonomy.
It has the potential to, \textit{e.g.,} enable a comprehensive description of the navigational risks or disambiguate challenging situations in planning or mapping.
For many of the currently employed robotic systems, the long-term stability of maps is a pertaining issue due to the often limited metrical understanding of the environment for which high-level semantic information is a possible solution. 

With the advances in deep learning, vision-based semantic segmentation frameworks have become a very mature field.
While there has also been significant progress on LiDAR-based semantic segmentation frameworks, it is still not as developed as their vision-based counterpart.

Nevertheless, LiDAR-based approaches have certain crucial advantages over other modalities as they are unaffected by the illumination conditions of the environment.
This is in contrast to cameras which provide crucial descriptive information but are heavily affected by poor lighting conditions.
Consequently, LiDAR-based systems effectively provide a more resilient segmentation system for a variety of challenging scenarios, such as operating at night and dynamically changing lighting conditions.

Many existing approaches operate using projection models, which typically transform the irregular pointcloud data into an ordered 2D domain, allowing them to utilize the extensive research available for images.
The downside is that this requires a predefined configuration based on the number of beams, angular resolution, and vertical \ac{FoV}.
LiDAR systems differ in these properties, which means that changing the sensory system after training might yield projections with geometrical and structural scarcity.
Consequently, the resulting projection is often insufficient to express the complexity of arbitrary environments accurately.
Accordingly, utilizing these approaches in generic environments with an arbitrary sensor system is often impossible without refining the initial network on the data from the new sensor environment.

LiDAR sensors are known to yield accurate geometrical and structural cues. 
Thus, modern LiDAR sensors often provide large \ac{FoV} measurements to precisely measure the robot's surroundings.
However, the projection onto the 2D domain of such large \ac{FoV} scans introduces distortions that deform the physical dimensions the environment.
Thus, dealing with different \ac{FoV}, sensor frequencies, and scales remains an open research problem for which the input representation constitutes a major factor.
\begin{figure}[!t]
  \centering
  \includegraphics[width=0.47\textwidth, trim={0.0cm, 0.6cm, 0.0cm, 0cm}, clip]{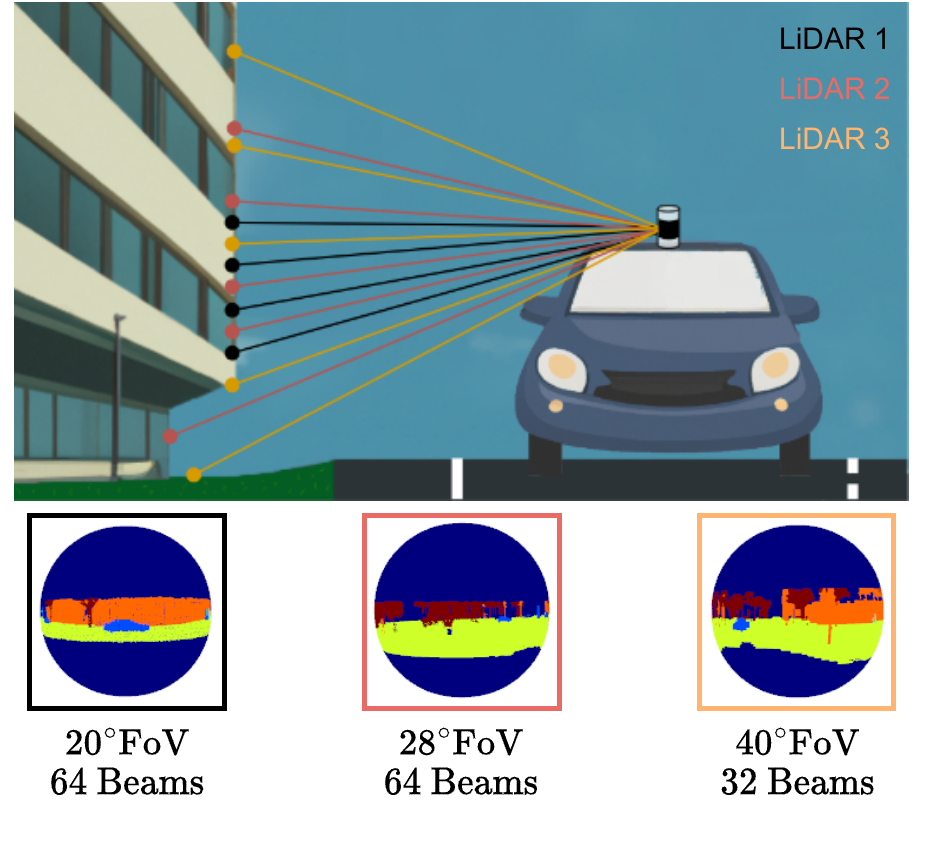}
  \caption{We propose a spherical semantic segmentation framework that can handle pointclouds from various LiDAR sensors with different vertical field-of-view and angular resolutions.}
  \label{figs:intro:teaser}
  \vspace{-0.75cm}
\end{figure}
In recent years, LiDAR sensors have become more affordable and available, becoming abundant in the context of robotics.
However, many state-of-the-art segmentation methods are often limited to a particular sensor and cannot benefit from multiple LiDARs due to their representation of the input.
For example, systems employing multiple LiDARs pointing in different directions are typically processed sequentially or using multiple instances of the same network, one for each sensor.
However, more crucial insights and structural dependencies in overlapping areas can be considered by jointly predicting the segmentation using all available sensors.


In this work, we propose a framework that takes LiDAR scans as input (cf. Figure~\ref{figs:intro:teaser}), projects them onto a sphere, and utilizes a spherical \ac{CNN} for the task of semantic segmentation.
The projection of the LiDAR scans onto the sphere does not introduce any distortions and is independent of the utilized LiDAR, thus, yielding an agnostic representation for various LiDAR systems with different vertical \ac{FoV}. 
We adapt the structure of common 2D encoder and decoder networks and support simultaneous training on different datasets obtained with varying LiDAR sensors and parameters without having to adapt our configuration.
Moreover, since our approach is invariant to rotations due to the spherical representation, we support arbitrarily rotated input pointclouds.
In summary, the key contributions of this paper are as follows:
\begin{itemize}
    \item A spherical end-to-end pipeline for semantic segmentation supporting various input configurations. 
    \item A spherical encoder-decoder structure including a spectral pooling and unpooling operation for  $SO(3)$ signals.
\end{itemize}
\section{Related Work}
\label{sec:related_work}
\begin{figure*}[!t]
  \centering
   \includegraphics[width=1.0\textwidth, trim={0.0cm, 0.2cm, 0.0cm, 0cm}, clip]{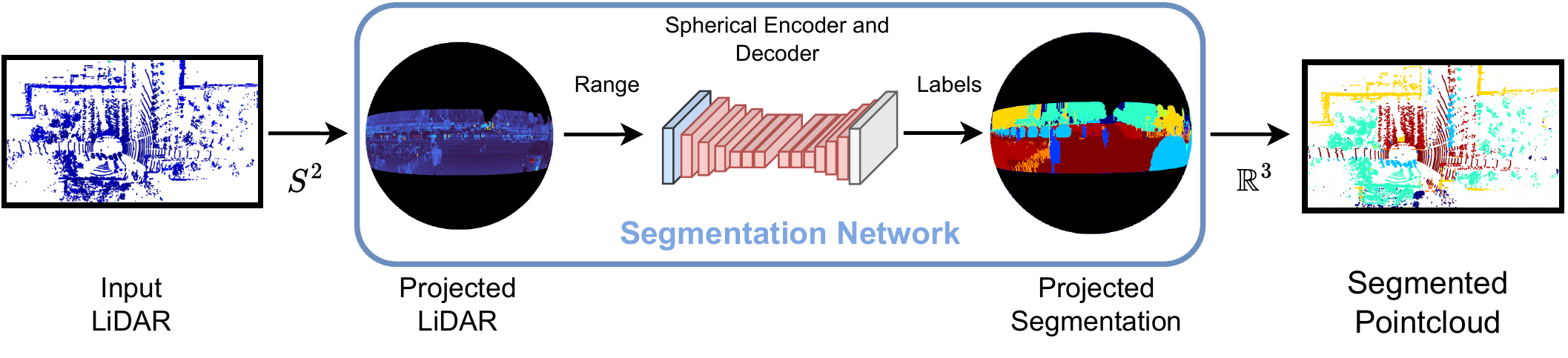}
   \caption{Overview of our proposed multi-modal segmentation framework. Initially, we employ a base network to a LiDAR scan to get a semantic segmentation. Next, the segmentation is fused with all camera images to get a refined solution. Finally, the spherical segmentation is back-projected into its original form.}
   \label{figs:method:overview}
   \vspace{-0.6cm}
\end{figure*}
Methods using a LiDAR have to deal with the inherent sparsity and irregularity of the data in contrast to vision-based approaches.
Moreover, LiDAR-based methods have various choices on how to represent the input data~\cite{gao2021we}, including, directly using the pointcloud~\cite{Charles2017a,Qi2017b},  voxel-based~\cite{Tchapmi2018,Zhou2020,Xu2022} or projection-based~\cite{Milioto2019a,Cortinhal2020, Wu2018, alonso20203d} representations.
The selection of the input representation, which yields the best performance for a specific task, however, still remains an open research question.

Direct methods such as PointNet~\cite{Charles2017a, Qi2017b} operate on the raw unordered pointcloud and extract local contextual features using point convolutions~\cite{Li2018a}. 
Voxel-based approaches~\cite{Tchapmi2018, Cai2016, Zhou2020} keep all the geometric understanding of the environment and can readily accumulate multiple scans either chronologically or from different sensors. 
SpSequenceNet~\cite{Shi2020} explicitly uses 4D pointclouds and considers the temporal information between consecutive scans. 
However, it is evident that the computational complexity of voxel-based approaches is high due to their high-dimensional convolutions, and their accuracy and performance are directly linked to the chosen voxel size, which resulted in works that organize the pointclouds into an Octree, Kdtree, etc.~\cite{Su2018} for efficiency.
Furthermore, instead of using a cartesian grid, PolarNet~\cite{Zhang_2020_CVPR} discretizes the space using a polar grid and shows superior quality. 
A different direction of research is offered by graph-based approaches~\cite{Wang2019a} which can seamlessly model the irregular structure of pointclouds though more experimental directions in terms of graph building, and network design are still to be addressed. 

Projection-based methods differ from other approaches by transforming the pointcloud into a specific domain, such as 2D images, which the majority of projection-based methods~\cite{Milioto2019a, Wu2018, Cortinhal2020, alonso20203d} rely on.

Furthermore, projections to 2D images are appealing as it enables leveraging all the research in image-based deep learning but generally need to rely on the limited amount of labeled pointcloud data.
Hence, the work of Wu et al.~\cite{Wu2019} tackles the deficiency in labeled pointcloud data by using domain-adaption between synthetic and real-world data.

The downsides of the projection onto the 2D domain are: i) the lack of a detailed geometric understanding of the environment and ii) the large \ac{FoV} of LiDARs, which produces significant distortions, decreasing the accuracy of these methods.
Hence, recent approaches have explored using a combination of several representations~\cite{Xu2022, Li2022} and convolutions~\cite{Park2022}.
Recent works~\cite{gerdzhev2021tornado, duerr2022rangebird} additionally learn and extract features from a Bird's Eye View projection that would otherwise be difficult to retain with a 2D projection.

In contrast to 2D image projections, projecting onto the sphere is a more suitable representation for such large \ac{FoV} sensors.
Recently, spherical \ac{CNN}s~\cite{Esteves2018, Cohen2018a, Esteves2020} have shown great potential for, \textit{e.g.,} omnidirectional images~\cite{Lohit2020, Bernreiter2021} and cortical surfaces~\cite{Zhao2019, Zhao2021}.

Moreover, Lohit et al.~\cite{Lohit2020} proposes an encoder-decoder spherical network design that is rotation-invariant by performing a global average pooling of the encoded feature map.
However, their work discards the rotation information of the input signals and thus, needs a special loss that includes a spherical correlation to find the unknown rotation w.r.t. the ground truth labels.

Considering the findings above, we propose a composition of spherical \ac{CNN}s, based on the work of Cohen et al.~\cite{Cohen2018a}, that semantically segments pointclouds from various LiDAR sensor configurations.
%

%
\section{Spherical Semantic Segmentation}
\label{sec:method}
This section describes the core modules of our spherical semantic segmentation framework, which mainly operates in three stages: i) feature projection, ii) semantic segmentation, and iii) back-projection (\textit{cf.} Figure~\ref{figs:method:overview}).

Initially, we discuss the projection of LiDAR pointclouds onto the unit sphere and the feature representation that serves as input to the spherical \ac{CNN}.
Next, we describe the details of our network design and architecture used to learn a semantic segmentation of LiDAR scans. 
\subsection{Sensor Projection and Feature Representation}\label{sec:method:feature}
Initially, the input to our spherical segmentation network is a signal defined on the sphere $S^2=\left\{\vec{p} \in \mathbb{R}^3 \, | \, \norm{\vec{p}}_2=1\right\}$, with the parametrization as proposed by Healy et al.~\cite{Healy2003}, \textit{i.e.} 
\begin{equation}
    \vec{\omega}(\phi,\theta)=[\cos\phi\sin\theta, \sin\phi\sin\theta,\cos\theta]^\top,
\end{equation}
where $\vec{\omega}\in S^2$, and $\phi\in[0,2\pi]$ and $\theta\in[0,\pi]$ are azimuthal and polar angle, respectively.

We then operate in an end-to-end fashion by transforming the input modality (\textit{i.e.,} the pointcloud scan) into a spherical representation in $S^2$. 
Consequently, as an initial step, a pointcloud is projected onto the unit sphere, \textit{e.g.} the $j$th point $\vec{p}_j=[x_j,y_j,z_j]^\top$ is projected using
\begin{align}\label{eq:method:proj}
    \phi_j=\arctan\left(\frac{y_j}{x_j}\right),\hspace{1cm}
    \theta_j=\arccos\left(\frac{z_j}{||\vec{p}_j||_2}\right),
\end{align}
From the LiDAR projection, we sample the range values using an equiangular grid complying with the sampling theorem by Discroll and Healy~\cite{Driscoll1994}.
It is important to note that we omit the sampling of the intensity/remission values of the LiDAR scans since their values significantly fluctuate between LiDAR sensors, particularly between different manufacturers.
Thus, would require additional investigation and further calibration of the input data.

Although in this work, we only utilize measurements from a LiDAR scanner, our approach is not limited to this sensor type. 
Other sensory systems, such as multi-camera systems, thermal cameras, and depth sensory systems, can seamlessly be integrated by performing the same spherical projection.
Additionally, since all sensors would share the same feature representation, our approach readily facilitates various combinations of heterogeneous sensory types~\cite{Bernreiter2021}.
Moreover, despite utilizing only LiDAR sensors defined over $360^\circ$, our sampling approach is agnostic to the resolution and field-of-view of the sensor and can be used with arbitrary viewpoint coverage.

Finally, after the projection and sampling, the spherical network will receive the LiDAR scan as a feature vector $\in\mathbb{R}^{2\times{}2\text{BW}\times{}2\text{BW}}$, where be $\text{BW}$ corresponds to the spherical bandwidth used for the equiangular sampling~\cite{Driscoll1994}.
The chosen bandwidth directly controls the employed spatial discretization and, consequently, the spectrum's resolution in the frequency domain.
In particular, the higher the bandwidth, the finer the spectral resolution, and more memory is required, which needs to be carefully considered when designing the spherical network, \textit{e.g.,} the first convolutional layer is designed to operate on the exact $\text{BW}$ as used by the sampling but later layers will decrease the bandwidth.
In the following, we discuss the design of the remaining layers in our network.
\subsection{Network Design and Architecture}
In this section, we will discuss the design choices of our network that takes LiDAR scans as input and is able to learn a semantic segmentation of it.
\begin{figure*}[!t]
  \centering
   \includegraphics[width=0.9\textwidth, trim={0.0cm, 0.1cm, 0.0cm, 0cm}, clip]{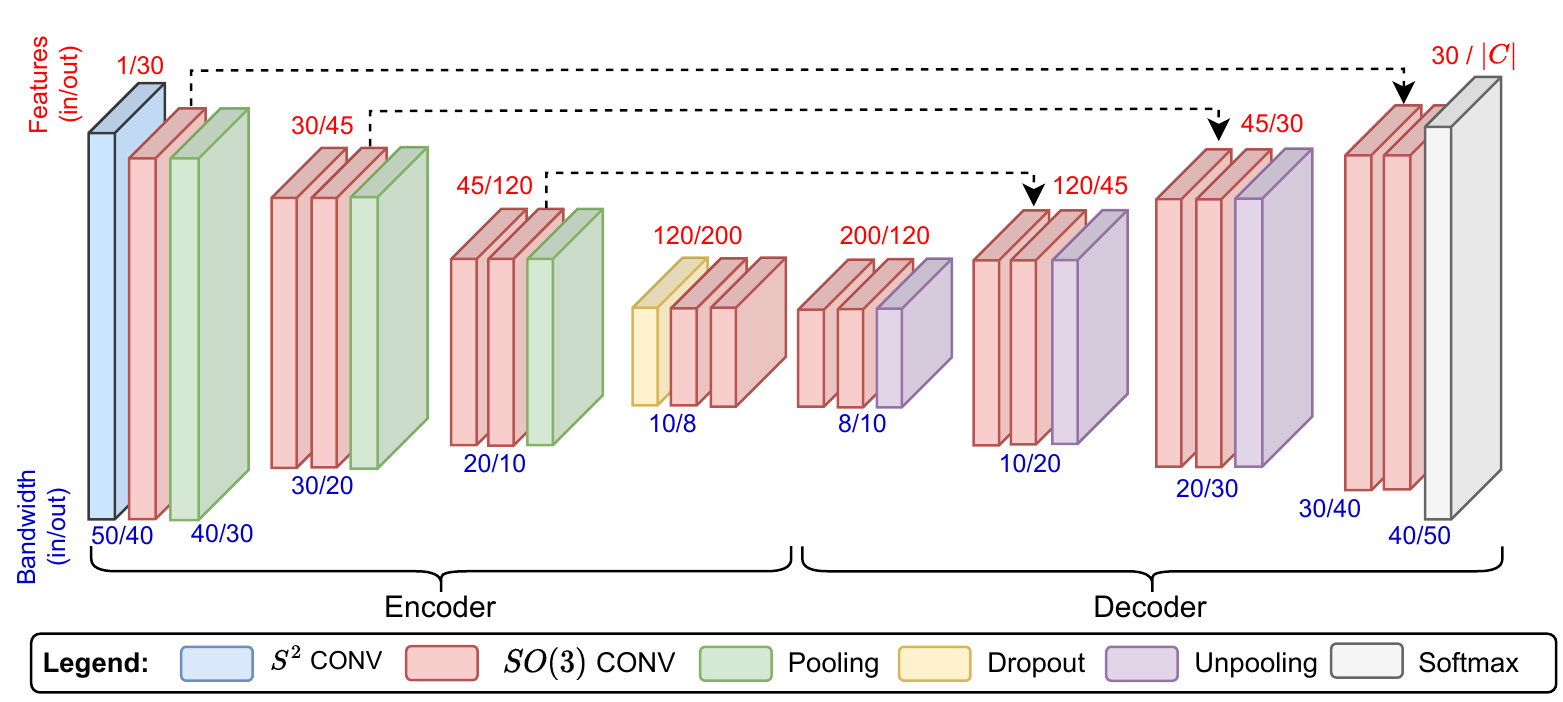}
   \caption{Proposed network architecture. We use an encoder-decoder design where the dotted lines denote skip connections. Initially, our network lifts the features to $SO(3)$ and will eventually integrate back to $S^2$ for the semantic segmentation. The network puts out the logits for the number of configured classes $C$, which then, together with a softmax, results in a semantic segmentation of the input pointcloud.}
   \label{figs:s2ae:method:network}
   \vspace{-0.6cm}
\end{figure*}
%
The input of the base network will be the sampled spherical features from the LiDAR pointcloud, i.e., the range values.
We found that an initial spherical bandwidth between $50$ - $120$ yields a good trade-off between accuracy and memory consumption.

Overall, the network design is based on a spherical encoder-decoder structure.
The spherical encoder network increases the number of features from layer to layer while the bandwidth is decreased instead.
In a similar vein, the spherical decoder decreases the number of features while increasing the bandwidth from layer to layer back to the originally utilized bandwidth.
An overview of our spherical network architecture is given in Figure~\ref{figs:s2ae:method:network}.
Moreover, by decreasing and increasing the bandwidth during encoding and decoding, we will also increase and decrease the size of the kernels for consecutive convolutions, respectively.
%
\subsubsection{Feature Encoding}
Our network is based on the work of Cohen et al. ~\cite{Cohen2018a}. 
Thus we lift the features to $SO(3)$ during the encoding and have to revert to $S^2$ during the decoding.
In other words, only the initial layer of the network performs a convolution over $S^2$, whereas the remaining convolutional layers act on $SO(3)$ to preserve the convolution's equivariance property~\cite{Cohen2017} (\textit{cf.} the left part in Figure~\ref{figs:s2ae:method:network}). 
During the convolutions, we employ spatially localized kernels that are rotated around the sphere using operations in $SO(3)$.
Here, $SO(3)$ denotes the three-dimensional rotation group consisting of $\alpha$, $\beta$, and $\gamma$, corresponding to roll, pitch, and yaw (RPY).

Each convolution in our network is efficiently implemented based on the convolution theorem by performing a Fourier transform in $S^2$ and $SO(3)$~\cite{Kostelec2008}, respectively, \textit{i.e.} the convolution between two signals $f$ and $g$ is given by
\begin{equation}\label{eq:s2ae:method:conv}
    f \ast g = \mathcal{F}^{-1}\left\{\mathcal{F}\left\{f\right\}\cdot\mathcal{F}\left\{g\right\}\right\},
\end{equation}
where $\mathcal{F}$ is either a $S^2$ or $SO(3)$ Fourier transform and $\mathcal{F}^{-1}$ its inverse.
Additionally, we apply a PReLu~\cite{He2015} activation function followed by a three-dimensional batch normalization after each convolution. 
The last convolutional block during the encoding also applies a dropout for additional regularization.

In our approach, the $S^2$ convolution increases the number of features but is not done in place, \textit{i.e.,} it does not preserve the input bandwidth. 
Rather, it directly decreases the output bandwidth as part of the $S^2$ Fourier transform by having a smaller output bandwidth for the inverse $S^2$ Fourier transform in Eq.~\eqref{eq:s2ae:method:conv}.
The $SO(3)$ convolution increases the number of input features but preserves the utilized spherical bandwidth. 
And, the $SO(3)$ blocks decrease the bandwidth by applying a pooling operation in $SO(3)$.

Pooling and unpooling are done in the spectral domain of $SO(3)$, which has the advantage of retaining the equivariance~\cite{Rippel2015,Esteves2018} as opposed to spatial pooling.
In practice, the $SO(3)$ pooling is implemented by transforming the signal to the spectral domain using an $SO(3)$ Fourier transform~\cite{Kostelec2008,Healy2003} and subsequent low-pass filtering of the resulting spectrum. 
The inverse $SO(3)$ Fourier transform then yields the pooled $SO(3)$ signal. 
Between the forward and backward passes, we temporarily store the input bandwidth and inverse the operation by zero-padding the spectrum to the original size. 
\begin{table*}
\setlength{\tabcolsep}{5pt}
\centering
\begin{tabular}{c|cccccc|c}
\multicolumn{8}{c}{\textbf{Segmentation Quality Comparison in Terms of mIoU / Acc}} \\
\midrule
\textbf{Datasets} & & {nuScenes~\cite{Caesar2020}} & {SemanticKITTI~\cite{Behley2019}} & {SemanticPOSS~\cite{pan2020semanticposs}} & {Waymo~\cite{Sun_2020_CVPR}} & {A2D2~\cite{geyer2020a2d2}} & {PC-Urban~\cite{ibrahim2021annotation}} \\
\textbf{vFoV / Beams} & & $40^\circ$ / 32 & $28^\circ$ / 64 & $23^\circ$ / 40 & $20^\circ$ / 64 & $30^\circ$ / 16 & $45^\circ$ / 64\\
\midrule
\textbf{Methods} & $\varnothing$ \textbf{mIoU / Acc} & mIoU / Acc & mIoU / Acc & mIoU / Acc & mIoU / Acc & mIoU / Acc & mIoU / Acc\\
SqueezeSeg~\cite{Wu2018} & 9.5 / 57.0 & 9.5 / 57.1  & 10.8 / 64.5 & 9.5 / 57.3 & 6.7 / 40.1 & 10.6 / 63.9 & 9.9 / 59.3 \\
SqueezeSegV2~\cite{Wu2019} & 39.1 / 81.7 & 49.4 / 89.4 & 45.8 / 88.6 & 39.3 / 80.7 & 37.1 / 75.0 & 36.1 / 88.0 & 26.8 / 68.4 \\
RangeNet++~\cite{Milioto2019a} & 38.0 / 80.4 & 49.1 / 88.4 & 45.1 / 88.6 & 37.9 / 78.7 & 34.3 / 72.0 & 35.6 / 87.6 & 26.2 / 66.8 \\
3D-MiniNet~\cite{alonso20203d} & 42.6 / 81.2 & 54.9 / 90.5 & 51.1 / 92.5 & 44.3 / 84.8 & 45.9 / 81.5 & 39.3 / 89.7 & 20.1 / 48.1 \\
SalsaNext~\cite{Cortinhal2020} & 44.1 / 82.1 & \textbf{61.8 / 92.4} & 50.6 / 91.8 & 42.6 / 82.7 & 38.6 / 74.6 & 43.7 / 90.8 & 27.3 / 60.3 \\
Ours & \textbf{49.0 / 97.2} & 55.2 / {95.4} & \textbf{52.7 / 96.9} & \textbf{51.2 / 96.8} & \textbf{47.1 / 98.4} & \textbf{46.0 / 98.8} & \textbf{41.8 / 97.0}\\ 
\end{tabular}
\caption{Comparison of the mean IoU (mIoU) and the accuracy (Acc) in the validation split for each dataset. All datasets were part of the training split except for the PC-Urban dataset, which also contains a new LiDAR sensor that was not seen yet by the networks. The leftmost data column depicts the average mIou And Acc over all datasets. All values are given as percentages $[\%]$.}
\label{tab:s2ae:exp:miou_train}
\vspace{-0.9cm}
\end{table*}
\subsubsection{Feature Decoding}
All the convolutions in the spherical decoding component are done in $SO(3)$ (\textit{cf.} the right part in Figure~\ref{figs:s2ae:method:network}).
Moreover, we unpool the input signals to increase the bandwidth to match the input bandwidth again.
In particular, we apply the $SO(3)$ Fourier transform to the input signal and zero-pad the resulting spectrum to a larger size.
Finally, the zero-padded signal is inverse $SO(3)$ Fourier transformed and passed to the next convolutional layer. 
Similar to the encoding part, we perform the inverse by applying an idealized low-pass filter for the backward pass.
The last convolution during decoding is different from the preceding operations as it zero-pads the convolved signal in the spectral domain to achieve the initial input bandwidth (sampling bandwidth).
Thus, the last layer does not rely on any unpooling operation.
Although the output has the correct size, it still needs to be mapped back to its original space, \textit{i.e.,} $S^2$. 
To transform the $SO(3)$ signal back to $S^2$, a max pooling operation or an integration of the $SO(3)$ signal over the last entry $\gamma$ (\textit{i.e.,} the yaw angle) can be used, resulting in $\mathbb{R}^{2\text{BW}\times2\text{BW}\times2\text{BW}}\mapsto\mathbb{R}^{2\text{BW}\times2\text{BW}}$. 
We have selected the latter approach for its efficient computation and simplicity.
During the inference, the spherical semantic segmentation is then achieved by applying a final softmax layer to the result of the integration. 
\subsubsection{Loss}
Finally, our proposed spherical network uses a common loss definition for semantic segmentation~\cite{Cortinhal2020, gerdzhev2021tornado}, \textit{i.e.} for prediction $\hat{y}$ and ground truth $y$ labels
\begin{align}\label{eq:method:loss}
    \mathcal{L}_{XC}(y,\hat{y}) &= -\sum_i w_i P(y_i) \mathrm{log}P(\hat{y}_i) \\
    \mathcal{L}_{LZ}(y,\hat{y}) &= \frac{1}{|C|}\sum_{c\in{}C} J(e(c))\\
    \mathcal{L}(y,\hat{y}) &= \mathcal{L}_{XC}(y,\hat{y}) + \mathcal{L}_{LZ}(y,\hat{y}),
\end{align}
where $\mathcal{L}_{XC}$ is a weighted cross-entropy loss where $w_i$ are the class weights and $P(\cdot)$ the corresponding probabilities. 
The latter term $\mathcal{L}_{LZ}$ is a lovasz-softmax~\cite{Berman2018} loss where J is the lovasz intersection over union, $e(c)$ the errors for class $c$ and $C$ the set of all classes.
Notably, the loss operates on the equiangular samples in $S^2$, and since we do not discard the rotational information of the signals on the sphere, we have a direct mapping between the input and the output signals.
\subsubsection{Back-Projection}
The final pointcloud segmentation in $\mathbb{R}^3$ is achieved by back-projecting the spherical projection from $S^2$ to $\mathbb{R}^3$ using the sampled range values, \textit{i.e.,} inverting Eq.~\eqref{eq:method:proj}, \textit{s.t.} for the $j$th projected point with $\phi_j$ and $\theta_j$
\begin{align}\label{eq:method:back_proj}
    x_j &= r_j \cdot \mathrm{cos}\left(\phi_j\right) \mathrm{sin}\left(\theta_j\right) \\
    y_j &= r_j \cdot \mathrm{sin}\left(\phi_j\right) \mathrm{sin}\left(\theta_j\right) \\
    z_j &= r_j \cdot \mathrm{cos}\left(\theta_j\right),
\end{align}
where $r_j = \norm{\vec{p}_j}_{2}$.
\section{Experiments}
\label{sec:experiments}
This section presents the experimental validation of our spherical segmentation framework, where we show that our pipeline gives an accurate semantic segmentation of the environment and that it generalizes well to different sensory setups.
We first validate the segmentation quality of our spherical network and compare it to current state-of-the-art projection-based segmentation frameworks.
Next, we demonstrate the flexibility of our representation by increasing the input field-of-view drastically.
Finally, we evaluate the computational cost to show the method's applicability to real-world scenarios.
We use the structure depicted in Figure~\ref{figs:s2ae:method:network} for all experiments.
\subsection{Experiment and Training Setup}\label{sec:s2ae:exp:base}
We validate our proposed approach by comparing the performance of our spherical network to state-of-the-art projection-based segmentation frameworks, RangeNet++~\cite{Milioto2019a}, SqueezeSeg~\cite{Wu2018, Wu2019}, 3D-MiniNet~\cite{alonso20203d}, and SalsaNext~\cite{Cortinhal2020}.

For the comparison, we utilize several datasets with various different LiDARs to show the ease of use of our proposed approach given sensors with different vertical \ac{FoV} (vFoV) and number of beams.
All networks use only the range values as feature and are trained on the nuScenes~\cite{Caesar2020}, SemanticKITTI~\cite{Behley2019}, SemanticPOSS~\cite{pan2020semanticposs}, Waymo~\cite{Sun_2020_CVPR}, A2D2~\cite{geyer2020a2d2}. 
We trained all networks for 50 epochs and used the \ac{mIoU} and accuracy metric to assess the quality of the similarity to the ground truth.

Moreover, since all these datasets provide significantly different classes, \textit{e.g.} nuScenes groups all objects into man-made objects.
Similar to Sanchez et al.~\cite{sanchez2022cola}, we abstracted the semantic classes to a total of five classes in order to provide a shared set of classes between the datasets: vehicles, persons, ground, man-made, and vegetation.

In addition, for each dataset that provides a split between training and validation, we randomly selected 4000 pointclouds from the former for training the networks and used the latter for inference.
For the datasets without a split, we created a split and followed the above procedure for training and inference.

The input data to the projection-based methods was projected into the largest \ac{FoV} of the LiDARs known during training.
It is important to note that our approach, in contrast to the other ones used, does not require any such considerations and projects all pointclouds directly onto the sphere.
\begin{table}
\setlength{\tabcolsep}{5pt}
\centering
\begin{tabular}{c|c|c|c|c|c|c}
\multicolumn{7}{c}{\textbf{Segmentation Quality Comparison on a new Domain}} \\
\midrule
\textbf{Methods} & \textbf{mIoU} & \rotHeader{Vehicle} & \rotHeader{Person} & \rotHeader{Ground} & \rotHeader{Man-Made} & \rotHeader{Vegetation} \\
\midrule
SqueezeSeg~\cite{Wu2018} & 12.3 & 0.0 & 0.0 & 61.4 & 0.0 & 0.0 \\
SqueezeSegV2~\cite{Wu2019} & 18.2 & 22.2 & 3.1 & 58.0 & 7.6 & 0.0 \\
RangeNet++~\cite{Milioto2019a} & 21.5 & 22.6 & 4.7 & 60.4 & 20.0 & 0.0 \\
3D-MiniNet~\cite{alonso20203d} & 20.8 & 24.4 & \textbf{16.7} & 60.0 & 2.8 & 0.0 \\
SalsaNext~\cite{Cortinhal2020} & 22.7 & \textbf{32.7} & 9.5 & \textbf{62.8} & 8.3 & 0.0 \\
Ours & \textbf{30.1} & 19.5 & 1.8 & 59.9 & \textbf{23.7} & \textbf{49.3} \\ 
\end{tabular}
\caption{Comparison of the class-wise and mean IoU for the SemanticUSL~\cite{jiang2021lidarnet} dataset, which uses a different environment and sensory system than what was seen during training. All IoU values are given as percentages $[\%]$.}
\label{tab:s2ae:exp:semantic_usl}
\vspace{-1.5cm}
\end{table}
\subsection{Segmentation Quality Comparison and Validation}
We first evaluate the performance within the training domain, \textit{i.e.,} the training and test are in the same urban setting. The results are shown in Table~\ref{tab:s2ae:exp:miou_train}. 
One can see that most of the other baseline methods suffer from the large variations in the data and consequently also fluctuate in their performance significantly.
Notably, due to the difference in vFoV the projections distort the physical dimensions, resulting in less accuracy where vFov is smaller (\textit{e.g.,} SemanticPOSS~\cite{pan2020semanticposs} and Waymo~\cite{Sun_2020_CVPR}). 

In contrast, our proposed approach achieves a good segmentation of the input pointcloud and, most importantly, achieves consistent scores across the various datasets.
Although our approach does not achieve the segmentation quality for the nuScenes~\cite{Caesar2020} dataset in our tests, its primary benefit is the generalization capabilities between sensory systems. 
Thanks to the spherical representation of pointcloud data employed by our method, we can correctly perform the projection required by the different opening angles of the LiDARs. 
Consequently, our approach is less affected by the changing angular resolution between the datasets in comparison to the image-based projection approaches.

We additionally evaluate the networks on an unseen sensory system, PC-Urban~\cite{ibrahim2021annotation} to assess our approach's generalization capabilities within the same domain. 
The difference in vFoV ($12.5^\circ$ down and $-7.5^\circ$ up) results in warped physical proportions for the 2D projection-based methods, which greatly decreases their segmentation performance. 
Consequently, the segmentation quality of high objects such as man-made buildings and vegetation significantly degrades.  
Our approach is less affected by the change in the vFoV and hence, also attains the highest mIoU on this dataset.
\subsection{Segmentation Quality on a Different Domain} 
In this experiment, we test all approaches on the SemanticUSL~\cite{jiang2021lidarnet} ($45^\circ$ vFoV and 64 beams) dataset, which offers a different sensory system and environment (off-road and campus scenes) than what was seen by the networks during training. 
The change in the vFoV is the same as for the PC-Urban dataset in the previous experiment.
Table~\ref{tab:s2ae:exp:semantic_usl} shows a comparison between the class-wise and mean IoU.

Due to the vastly different environment and sensor intrinsics, all but our proposed method fail entirely to segment the vegetation class as the learned representation no longer matches.
Our approach is not affected by such warping of the physical dimensions and, therefore, achieves a better segmentation quality. 

However, it is difficult for our representation to disambiguate persons that are far away from man-made objects and small vegetation when only range values are available.
Providing additional information through multiple modalities is a possible solution and is left for future work. 
Nevertheless, our approach still maintains the highest mIoU compared to the other approaches.
This highlights the main benefit of our method, which is its ability to generalize and consequently provide an improved overall segmentation. 
\subsection{Semantic Segmentation of Rotated Pointclouds}\label{sec:s2ae:results:rotated}
Next, we show that our representation has the advantage of rotational invariance to the input data, allowing pointclouds to be arbitrarily rotated. 
This allows our method to support various input configurations such as different angular resolutions and tilted sensor mounts.

\begin{figure}[!thb]
  \centering
   \includegraphics[width=0.48\textwidth, trim={0.0cm, 0.1cm, 0.0cm, 0cm}, clip]{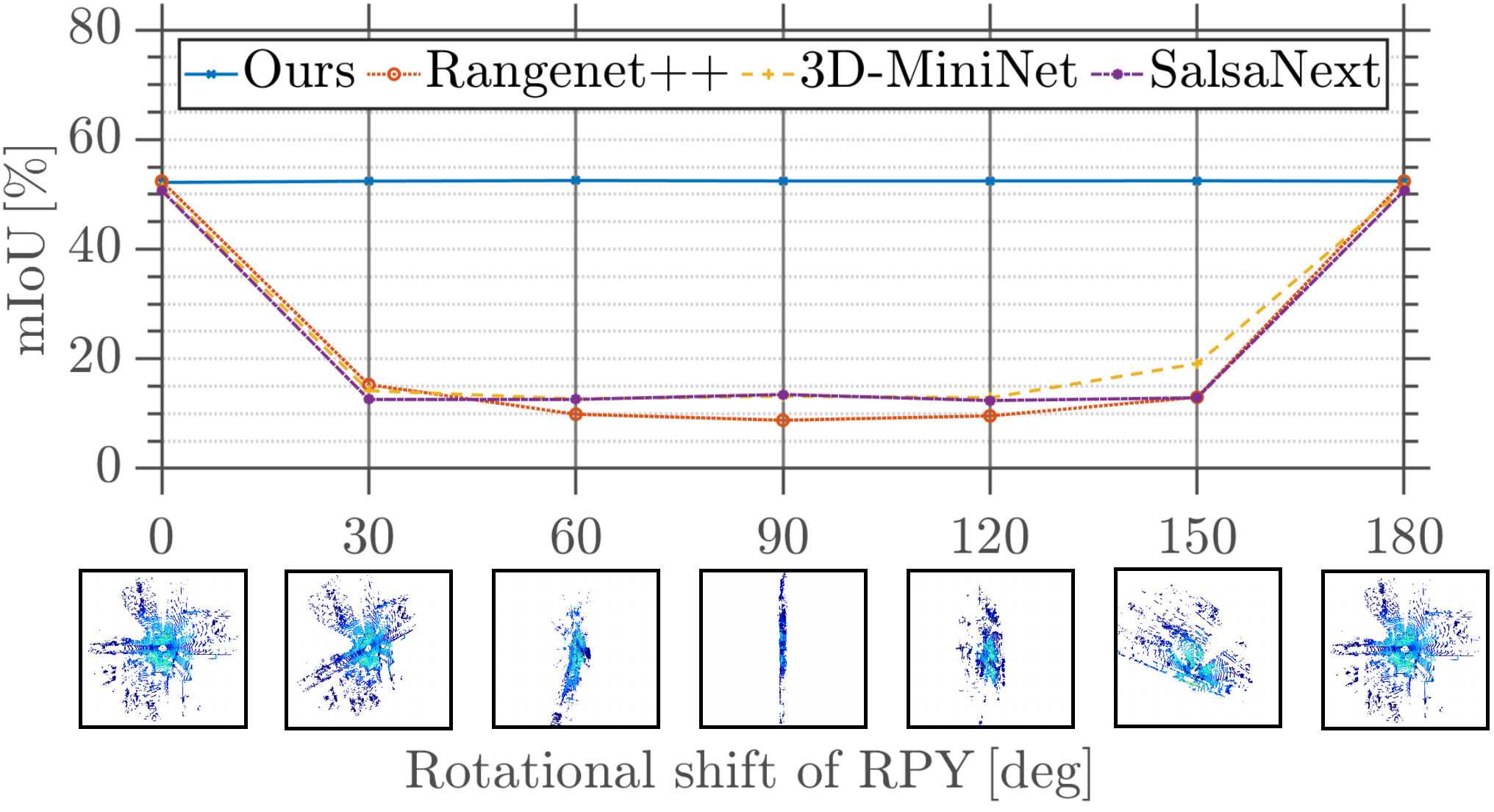}
   \caption{Comparison of the mIoU performance using different rotational shifts on the SemanticKITTI sequence 08. The bottom row of pointclouds shows exemplary results after applying the rotations.}
   \label{figs:s2ae:results:rotation}
   \vspace{-0.3cm}
\end{figure}

In this experiment, we applied a predefined rotation from $0^\circ$ to $180^\circ$ around the RPY axes of the input pointcloud.
Figure~\ref{figs:s2ae:results:rotation} shows the mIoU for various rotational shifts using sequence $08$ of the SemanticKITTI dataset, and the model trained in Section~\ref{sec:s2ae:exp:base}.
The rotation of the pointclouds yields odd and ineffective representations of the scans, and thus, all other projection-based methods experience a large drop in their performance.
Note that since the rotation of $180^\circ$ around RPY results in the original pointcloud, the initial mIoU is restored again.
Generally, 2D projection methods implicitly require the pointclouds to be horizontally oriented for an efficient prediction. 
Hence, it is particularly difficult for these methods to utilize multiple LiDARs simultaneously if one of the sensors is tilted w.r.t. the other ones.

The spherical projection is not only more efficient and natural, but the spherical Fourier transform is also invariant to rotations. 
Thus, our approach is completely unaffected by the rotated pointclouds and maintains the mIoU over all rotational shifts.
\subsection{Runtime Evaluation}
\begin{figure}[!thb]
  \centering
   \includegraphics[width=0.42\textwidth, trim={0.0cm, 0.1cm, 0.0cm, 0cm}, clip]{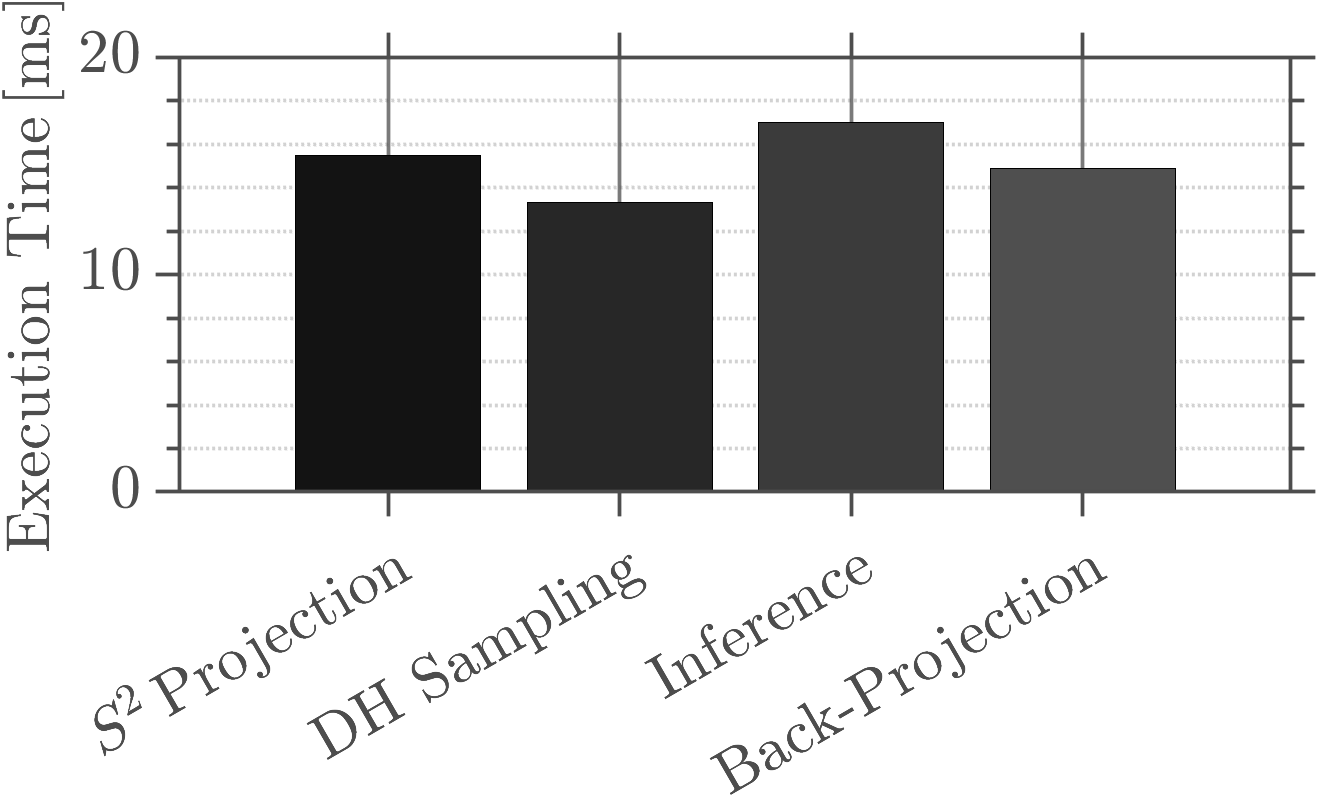}
   \caption{Execution time in $\mathrm{ms}$ partitioned per component. All values are averaged over 1000 samples. Our proposed approach is able to segment an input pointcloud in approximately $60\,\mathrm{ms}$.}
   \label{figs:s2ae:results:performance}
   \vspace{-0.3cm}
\end{figure}

Finally, we assess the runtime performance of our proposed approach to understand its potential for deploying it in real-world applications. 
Figure~\ref{figs:s2ae:results:performance} shows the evaluation of the execution time for the different components of our proposed approach. 
In this experiment, we consider a sampling bandwidth of $50$ for the LiDAR as depicted in Figure~\ref{figs:s2ae:method:network}.
The benchmark was performed on an Intel Xeon E5-2640v3 with an NVIDIA Titan RTX, and all scripts are written using PyTorch. 
It is evident that the discretization of the irregular pointcloud data on the sphere takes a considerable portion.
Nevertheless, our approach is able to infer a semantic segmentation in approximately $60\,\mathrm{ms}$.

\section{Conclusion and Future Work}\label{sec:summary}
In this paper, we proposed a spherical representation of pointclouds that can be used to train a model using various LiDAR sensors with different parameters. 
We presented in this context an end-to-end approach for semantic segmentation based on a spherical encoder-decoder network and showed that the spherical representation is a much more favorable representation, especially for high \ac{FoV} LiDAR systems.
Most importantly, our findings also indicate that our approach is invariant to rotations and has a better generalization to unseen LiDAR systems after training a model. 
Furthermore, our proposed approach is not limited to depth sensors, and other sensor types, such as RGB and thermal cameras, can be readily incorporated~\cite{Bernreiter2021}.

In future research, we intend to investigate two separate research directions. 
First, we explore the fusion of multiple camera images to refine the initial semantic segmentation and its ontology.
Second, we intend to migrate our spherical network to be fully in $S^2$ in order to decrease the memory requirements and improve its practical applicability.
\addtolength{\textheight}{-1cm} 
%
%
%
\bibliographystyle{IEEEtran}
\bibliography{bib/references.bib}

\end{document}